\pdfoutput=1 % may be required for arXiv PDFLaTeX compilation

\documentclass[conference]{IEEEtran}

\usepackage{cite}
\usepackage{amsmath,amssymb,amsfonts}
\usepackage{algorithmic}
\usepackage{graphicx}
\usepackage{textcomp}
\usepackage{xcolor}

\newcommand{\Vc}[1]{{\bf #1}}
\newcommand{\Vg}[1]{\boldsymbol{#1}} % for bold greek letters

\newcommand{\Trn}[1]{{\bf #1}^{\mbox{\sf \tiny T}}}
\newcommand{\NTr}[1]{{#1}^{\mbox{\sf \tiny T}}}

\newcommand{\Real}{\mathbb{R}}

\newcommand{\DE}{\stackrel{\text{def}}{=}}

\newcommand{\NExFrac}[2]{\exp\!\left(-\frac{#1}{#2}\right)}

\begin{document}
%>>>>>>>>>>>>>>>>>>>>>>>>>>>>>>>>>>>>>>>>>>>>>>>>>>>>>>>>>>>>>>>>>>>>>>>>>>>>>>>>>>>>>>>>>>>>>>>>>>>>>>>>>>>>>>>>>>>>>>>

\title{Compression of 3D Gaussian Splatting Data Using GPU-friendly Graphics Texture Coding}

\author{\IEEEauthorblockN{Amir Said}
\IEEEauthorblockA{\textit{Qualcomm AI Research}\\
San Diego, CA, USA \\
asaid@qti.qualcomm.com, said@ieee.org}
\and
\IEEEauthorblockN{Randall Rauwendaal}
\IEEEauthorblockA{\textit{Qualcomm Graphics Research}\\
Boulder, CO, USA \\
rrauwend@qti.qualcomm.com}
}

\maketitle

%>>>>>>>>>>>>>>>>>>>>>>>>>>>>>>>>>>>>>>>>>>>>>>>>>>>>>>>>>>>>>>>>>>>>>>>>>>>>>>>>>>>>>>>>>>>>>>>>>>>>>>>>>>>>>>>>>>>>>>>

\begin{abstract}
Techniques for modeling 3D scenes from image collections, such as 3D Gaussian Splatting (3DGS), are capable of generating high-quality novel views by leveraging graphics primitives with view-dependent appearance. In 3DGS, spherical harmonic (SH) are employed to model view-dependent color, resulting in a large number of SH coefficients per primitive and large memory requirements. While compression approaches have been proposed to mitigate this problem, they do not exploit the capabilities of modern Graphics Processing Units (GPUs) for parallel decoding and rendering. In this paper, we propose a method for compressing SH color coefficients using texture compression schemes specifically designed for efficient parallel GPU decoding and supported by dedicated hardware acceleration. It is shown that those methods can compress color coefficients more effectively than 2D textures by exploiting the fact that primitives can be locally grouped and reordered according to color. Furthermore, we introduce a bit-rate control strategy that preserves random access, enabling large-scale parallelization without compromising rendering performance. Experimental results using BC1 and BC7 texture compression formats show that GPU-based decompression can be achieved with negligible or imperceptible degradation in the visual quality of rendered 3DGS scenes.
\end{abstract}


%>>>>>>>>>>>>>>>>>>>>>>>>>>>>>>>>>>>>>>>>>>>>>>>>>>>>>>>>>>>>>>>>>>>>>>>>>>>>>>>>>>>>>>>>>>>>>>>>>>>>>>>>>>>>>>>>>>>>>>>

%>>>>>>>>>>>>>>>>>>>>>>>>>>>>>>>>>>>>>>>>>>>>>>>>>>>>>>>>>>>>>>>>>>>>>>>>>>>>>>>>>>>>>>>>>>>>>>>>>>>>>>>>>>>>>>>>>>>>>>>
\section{Introduction}\label{sc:Intro}
%>>>>>>>>>>>>>>>>>>>>>>>>>>>>>>>>>>>>>>>>>>>>>>>>>>>>>>>>>>>>>>>>>>>>>>>>>>>>>>>>>>>>>>>>>>>>>>>>>>>>>>>>>>>>>>>>>>>>>>>

\begin{figure}
\centerline{\includegraphics[scale=0.7]{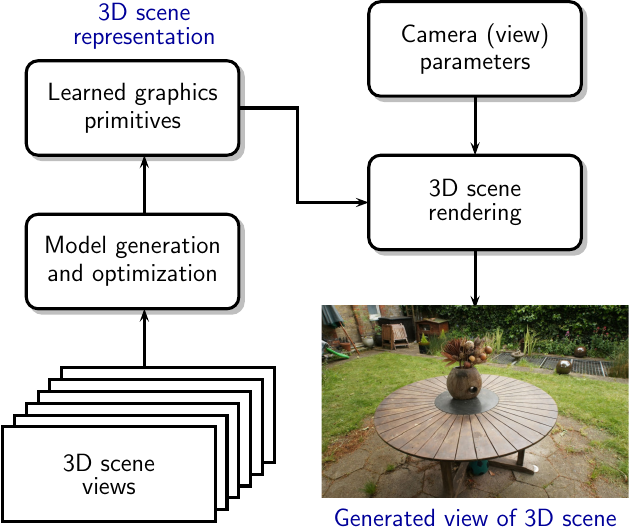}}
\caption{3D scene modeling and novel view synthesis using learned graphics primitives\label{fg:ModelView}.}
\end{figure}

Novel view synthesis models of natural 3D scenes can be created using the scheme shown in Fig.~\ref{fg:ModelView}, where a collection of scene views (images) is used to create a set of learned graphics primitives, which have their parameters optimized to minimize the visual differences between rendered and reference views. From those learned scene models the new views are generated by rendering the model's learned graphics primitives.

While this approach has been studied for decades, it only started to be widely adopted with the introduction of 3D Gaussian Splatting (3DGS)~\cite{Kerbl:23:3dg}, which creates scene models composed of learned 3D geometric primitives with a Gaussian spatial distribution of color and opacity, and visualized with the volumetric rendering splatting technique~\cite{zwicker:02:ewa}. 

There has been a significant amount of research for improving and extending 3DGS~\cite{Tong:24:rad,Fei:25:gss,Bao:25:gss,Chen:26:sur}, and it has been demonstrated that, with enough high-resolution views for learning, 3DGS views could be rendered with relatively low computational complexity, and visual quality indistinguishable from photos (for instance, the image shown in Fig.~\ref{fg:ModelView} was generated with 3DGS). 

% ·  ·  ·  ·  ·  ·  ·  ·  ·  ·  ·  ·  ·  ·  ·  ·  ·  ·  ·  ·  ·  ·  ·  ·  ·  ·  ·  ·  ·  ·  ·  ·  ·  ·  ·  ·  ·  ·  ·  ·
% Learned 3D models for novel view synthesis 
%\begin{figure}[htbp]
% ·  ·  ·  ·  ·  ·  ·  ·  ·  ·  ·  ·  ·  ·  ·  ·  ·  ·  ·  ·  ·  ·  ·  ·  ·  ·  ·  ·  ·  ·  ·  ·  ·  ·  ·  ·  ·  ·  ·  ·

One practical problem of learned 3D scene models is that they can only generate high-quality views of scenes containing surfaces with complex appearance, like reflective or transparent materials, if the appearance of graphics primitives can vary according to view direction. For instance, 3DGS learns 16~RGB color vector coefficients and combine them using a spherical harmonic representation to provide view-dependent appearances (color) to its Gaussian primitives.

Furthermore, a large number of primitives is needed to obtain realistic scene views, since they must capture all details in textured and complex objects. The large number of primitives, combined with the number of parameters per primitive, result in learned models that require large amounts of memory, increasing storage, transmission and rendering costs. 

This strongly motivates the use of compression for reducing a model's memory size, and many compression techniques have been proposed for 3DGS data, showing significant data reduction~\cite{Bagdasarian:25:git,Bagdasarian:25:scm,Ali:26:cmp}. However, those solutions have one important practical limitation: they do not exploit the fact that 3DGS is a form of computer graphics, and that views are expected to be rendered using Graphics Processing Units (GPUs) that can only achieve maximum efficiency with wide scale parallelization.

\begin{figure}
\centerline{\includegraphics[scale=0.7]{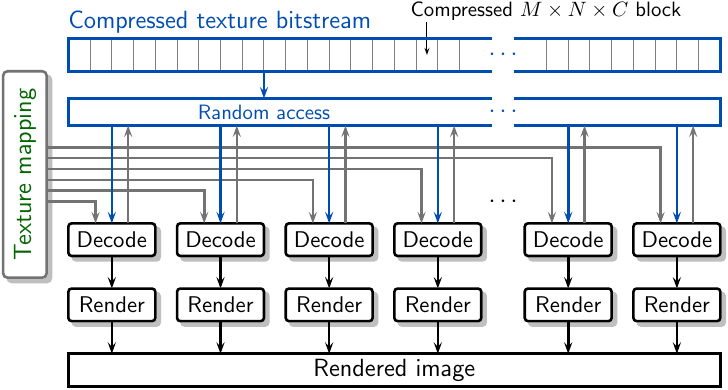}}
\caption{Parallel joint decoding and rendering of graphics textures within GPUs, enabled by random access to compressed data.\label{fg:TextRend}}
\end{figure}

We can find good examples of compression matched to GPU capabilities in the methods for compressing graphics textures~\cite{Nystad:12:ast,MSoft:25:tbc} that have been designed to compress blocks of texture elements ({\em texels}) with a fixed number of bits. This allows easy random access to a block's compressed data, since it can be located in a bitstream by simply multiplying the block index with the number of bits per block. This random access feature, in turn, enables very efficient forms of parallel execution of texture decoding and rendering, as shown in the diagram of Fig.~\ref{fg:TextRend}.

Thanks to this GPU parallel execution it is possible to render in real time highly complex scenes, comprised of billions of graphics primitives, because
\begin{itemize}
  \item Information in textures that are not visible in the current view are never decompressed.
	\item Within the compressed data, only the information needed for rendering is decompressed locally by the GPU cores.
	\item Maintaining the data in compressed form until actually needed reduces the bandwidth between texture memory and rendering cores, which is a major performance limiting factor.
  \item Texture decoding matches the GPU's parallelization architecture, scaling in each GPU generation.
	\item Custom GPU hardware acceleration for texture decompression further minimizes texture access times and energy requirements.
\end{itemize}

% ·  ·  ·  ·  ·  ·  ·  ·  ·  ·  ·  ·  ·  ·  ·  ·  ·  ·  ·  ·  ·  ·  ·  ·  ·  ·  ·  ·  ·  ·  ·  ·  ·  ·  ·  ·  ·  ·  ·  ·
% Parallel texture decoding and rendering
%\begin{figure}[htbp]
% ·  ·  ·  ·  ·  ·  ·  ·  ·  ·  ·  ·  ·  ·  ·  ·  ·  ·  ·  ·  ·  ·  ·  ·  ·  ·  ·  ·  ·  ·  ·  ·  ·  ·  ·  ·  ·  ·  ·  ·

In this paper we propose methods for compressing 3DGS data that employ methods developed for standard texture compression, as a form to extend all the advantageous features listed above to 3DGS rendering, and facilitate rendering very complex and large scenes.

The paper also addresses one potential limitation of this approach, which is the fact that it must employ forms of compression that supports random access to compressed data. In information theory terms, random access requires encoding data with a constant bitrate while trying to minimize the distortion (reproduction errors).

The consequence of this very strong constraint is that texture compression methods are significantly less effective than image coding methods that support variable bitrates (like JPEG), i.e., compression methods that allow varying the number of bits employed for each block according to its data characteristics.

This problem cannot be avoided for graphics texture compression because texels are encoded together in blocks within two-dimensional grids, which cannot be modified because they are required by texture mapping during rendering.

The parameters of 3DGS primitives, on the other hand, can be reorganized. In fact, the splatting and opacity blending processes require resorting primitives according to distance to camera before rendering each view~\cite{Kerbl:23:3dg}. This freedom to organize and group 3DGS data before coding has been exploited to improve conventional compression. 

In this paper we show that it can be leveraged to greatly improve constant bitrate compression, while preserving the ability to randomly access compressed data. We study the following two approaches
\begin{itemize}
  \item It is demonstrated that partially sorting primitives according to color can result in significantly lower distortion.
	\item Using multiple compressed data bitstreams, different forms of fixed-bitrate coding can be applied to primitives with different number of color components for view-dependent appearance, enabling more efficient variable assignments of bitrates.
\end{itemize}

In the following sections we define how to employ standard graphics texture coding methods to 3DGS primitive parameters.

%On one hand, most primitives have appearance that can be modeled with a few parameters, but an important fraction needs a significantly larger number of parameters to appear realistic.

%<<<<<<<<<<<<<<<<<<<<<<<<<<<<<<<<<<<<<<<<<<<<<<<<<<<<<<<<<<<<<<<<<<<<<<<<<<<<<<<<<<<<<<<<<<<<<<<<<<<<<<<<<<<<<<<<<<<<<<<

%>>>>>>>>>>>>>>>>>>>>>>>>>>>>>>>>>>>>>>>>>>>>>>>>>>>>>>>>>>>>>>>>>>>>>>>>>>>>>>>>>>>>>>>>>>>>>>>>>>>>>>>>>>>>>>>>>>>>>>>
\section{3DGS primitive parameters}\label{sc:Param}
%>>>>>>>>>>>>>>>>>>>>>>>>>>>>>>>>>>>>>>>>>>>>>>>>>>>>>>>>>>>>>>>>>>>>>>>>>>>>>>>>>>>>>>>>>>>>>>>>>>>>>>>>>>>>>>>>>>>>>>>

Information about how 3DGS and its extensions are implemented can be found in references like~\cite{Tong:24:rad,Fei:25:gss,Bao:25:gss,Chen:26:sur}. In this section we review the 3DGS formulation, constraining it to the original 3DGS implementation~\cite{Kerbl:23:3dg}, since it is the most well-known, and the proposed techniques can be easily extended to most modifications and extensions.

During the learning stage, each 3DGS primitive is defined by the following parameters:
\begin{itemize}
  \item $t \in \Real$ -- value for defining primitive's opacity;
  \item $\Vc{p} \in \Real^3$ -- primitive's Gaussian center in scene space;
  \item $\Vc{q} \in \Real^4$ -- quaternion defining Gaussian rotation;
  \item $\Vc{s} \in \Real^3$ -- Gaussian scaling vector;
  \item $\Vc{H} \in \Real^{3 \times N_h}$ -- matrix with $3 \times N_h$ spherical harmonic coefficients, where each column corresponds to 3-dimensional RGB color vectors.
\end{itemize}

To increase convergence, learning can start with $N_h = 1$ and later $N_h$ is progressively increased to 16. Using a variable number of spherical harmonic coefficients $N_h$ per primitive in the compressed data is discussed in Section~\ref{sc:Adapt}.

One common assumption is that those are the parameters that must to be compressed for scene rendering. In reality, those parameters are appropriate during learning, but $t$, $\Vc{s}$ and $\Vc{q}$ need to be converted for rendering, and as explained below, matrix $\Vc{H}$ should also be modified for compression and rendering.

The primitive's intrinsic opacity is obtained from a sigmoid transformation
\begin{equation}
  \mu \DE \frac{1}{1+e^{-t}} \in [0, 1]. \label{eq:sigmoid}
\end{equation}
while $\Vc{s}$ and $\Vc{q}$ define the symmetric Gaussian covariance matrix
\begin{equation}
  \Vg{\Sigma} \DE \Vc{R}(\Vc{q}) \Vc{D}(\Vc{s}) \Trn{D}(\Vc{s}) \Trn{R}(\Vc{q}) \in \Real^{3 \times 3}, \label{eq:covar}
\end{equation}
where $\Vc{D}$ and $\Vc{R}$ are respectively scaling and rotation matrices.

\begin{figure}
\centerline{\includegraphics[scale=0.6]{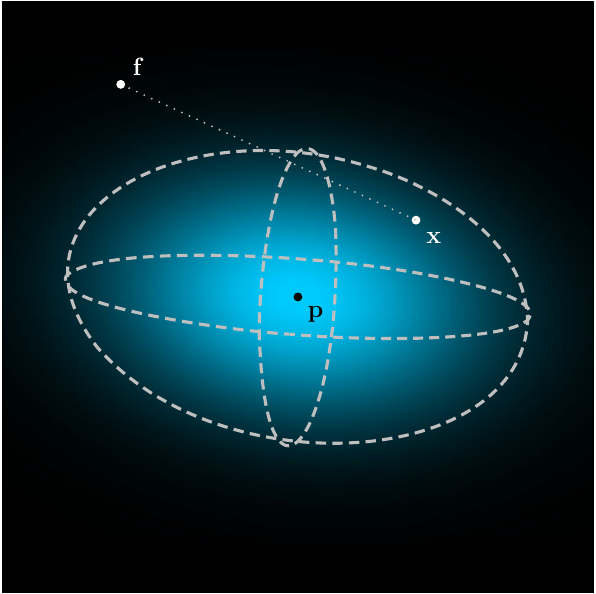}}
\caption{Rendering of a single 3D Gaussian Splatting primitive.\label{fg:Primtv}}
\end{figure}

Using those values, the primitive's opacity at position $\Vc{x}$ in space is
\begin{equation}
  \alpha(\Vc{x}) \DE \mu \, g(\Vc{x}) = \mu \, \NExFrac{\NTr{(\Vc{x} - \Vc{p})} \Vg{\Sigma}^{-1} (\Vc{x} - \Vc{p}) }{2}, \label{eq:opac}
\end{equation}
defining the primitive's ellipsoid shape, as illustrated in Fig.~\ref{fg:Primtv}.

% ·  ·  ·  ·  ·  ·  ·  ·  ·  ·  ·  ·  ·  ·  ·  ·  ·  ·  ·  ·  ·  ·  ·  ·  ·  ·  ·  ·  ·  ·  ·  ·  ·  ·  ·  ·  ·  ·  ·  ·
% 3DGS primitive as ellipsiod in space
%\begin{figure}[htbp]
% ·  ·  ·  ·  ·  ·  ·  ·  ·  ·  ·  ·  ·  ·  ·  ·  ·  ·  ·  ·  ·  ·  ·  ·  ·  ·  ·  ·  ·  ·  ·  ·  ·  ·  ·  ·  ·  ·  ·  ·

From a view position $\Vc{f}$, the primitive's color contribution at position $\Vc{x}$ (cf. Fig.~\ref{fg:Primtv}) can be defined as
\begin{equation}
  \Vc{c}(\Vc{f}) = \Vc{H} \, \Vc{w}(\Vc{f} - \Vc{p}),
\end{equation}
where $\Vc{w}(\cdot) \in \Real^{N_h}$ is the vector function for computing the spherical harmonic factors, defining the view-dependent color.

Using $i$ to represent the index of primitives sorted according to distance to camera, the view color is defined by the alpha blending equation
\begin{equation}
  \Vc{a} = \sum_{i=1}^{P} \alpha^{(i)} \Vc{c}^{(i)} \prod_{j=1}^{i-1} \left( 1 - \alpha^{(j)} \right). \label{eq:blend}
\end{equation}

Since the color vector $\Vc{c}$ is always multiplied by opacity $\alpha$, the alpha blending equation is equivalent to
\begin{equation}
  \Vc{a} = \sum_{i=1}^{P} \tilde{\Vc{c}}^{(i)} \prod_{j=1}^{i-1} \left( 1 - \alpha^{(j)} \right). \label{eq:preblend}
\end{equation}
where now
\begin{align}
  \tilde{\Vc{c}}(\Vc{x},\Vc{f}) & = \alpha(\Vc{x}) \, \Vc{H} \, \Vc{w}(\Vc{f} - \Vc{p}) \\
    & = g(\Vc{x}) \, \tilde{\Vc{H}} \, \Vc{w}(\Vc{f} - \Vc{p})
\end{align}
and
\begin{equation}
  \tilde{\Vc{H}} = \mu \, \Vc{H}. \label{eq:premult}
\end{equation}

This means we can use matrix $\tilde{\Vc{H}}$, with coefficients pre-multiplied by opacity $\mu$, directly as primitive parameter. This can save multiplications during rendering and, more importantly for compression, since $\mu \in [0,1]$, $\tilde{\Vc{H}}$ can be more efficiently encoded than $\Vc{H}$.

The parameters of each 3DGS primitive can be separated into the following categories:
\begin{itemize}
	\item {\bf Position:} vector $\Vc{p}$ with $(x, y, z)$ coordinates.
	\item {\bf Shape:} defined by the $3 \times 3$ matrix $\Vg{\Sigma}$. Since it is symmetric, only three coefficients need to be included in the compressed data.
	\item {\bf Opacity:} one parameter $\mu$ defining the opacity factor for each primitive. 
	\item {\bf Appearance:} matrix $\tilde{\Vc{H}}$ with $3 N_h$ elements.
\end{itemize}

Thus, the number of parameters per 3DGS primitive is $10 + 3 \times N_h$. In the case $N_h=16$ we have 58~parameters, meaning that without compression the spherical harmonic RGB coefficients represent 83\% of the total memory.

%<<<<<<<<<<<<<<<<<<<<<<<<<<<<<<<<<<<<<<<<<<<<<<<<<<<<<<<<<<<<<<<<<<<<<<<<<<<<<<<<<<<<<<<<<<<<<<<<<<<<<<<<<<<<<<<<<<<<<<<

%>>>>>>>>>>>>>>>>>>>>>>>>>>>>>>>>>>>>>>>>>>>>>>>>>>>>>>>>>>>>>>>>>>>>>>>>>>>>>>>>>>>>>>>>>>>>>>>>>>>>>>>>>>>>>>>>>>>>>>>
\section{Fixed bitrate compression}\label{sc:Compr}
%>>>>>>>>>>>>>>>>>>>>>>>>>>>>>>>>>>>>>>>>>>>>>>>>>>>>>>>>>>>>>>>>>>>>>>>>>>>>>>>>>>>>>>>>>>>>>>>>>>>>>>>>>>>>>>>>>>>>>>>

%=======================================================================================================================
\subsection{Block truncation coding}\label{sc:BTC}
%=======================================================================================================================

Image compression standards that use entropy coding, like JPEG, cannot efficiently provide random access with the fine granularity required during parallel rendering. Thus, they could not be used for graphics textures, and compression methods specialized for graphics textures and parallel decoding were developed.

The texture compression methods are designed to support the type of parallel implementation shown in Fig.~\ref{fg:TextRend}: texture blocks comprising of M×N×C texel data elements are compressed with a fixed number of bits (commonly bytes), enabling fast and simple random access to each block, which is decoded and used for rendering by each parallel processor.

In terms of data compression techniques, this corresponds to a type of vector quantization (VQ), However, conventional VQ techniques commonly employ large codebooks for decoding. This is relatively fast, but the codebooks typically are very large, making this approach unsuitable for parallel decoding, since all processors may need to keep reading large amounts of codebook data.

This vector quantization problem was solved for texture compression by using a type of image compression, called Block Truncation Coding (BTC)~\cite{Delp:79:btc,kurita:93:btc,hong:04:frb}, that generates a fixed number of bits per image block without using any additional codebook memory.

The most basic form of BTC for RGB color images is a form of vector quantization that uses $Q$ bits to store the indexes of quantized RGB vectors, with
\begin{equation}
  F =  2^Q - 1.
\end{equation}

\begin{figure}
\centerline{\includegraphics[scale=0.6]{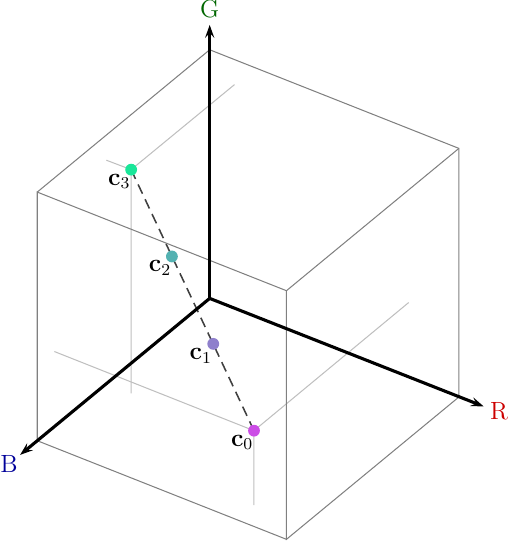}}
\caption{Example of how Block Truncation Coding (BTC) defines reproduction vectors and interpolation in RGB color space ($Q=2$~bits)\label{fg:BTCDiag}.}
\end{figure}

For each image block with $M \times N$ RGB pixels, two reference RGB vectors $(\Vc{c}_0, \Vc{c}_F)$ are chosen and encoded with $P$ bits used for each reference. Those BTC reference vectors define a sequence ${\cal L} = \{\Vc{c}_k\}_{k=0}^{F}$ of $2^Q$ equally spaced points in the RGB space, between points $c_0$ and $c_F$, i.e., for $k = 0, 1, \ldots, F$
\begin{equation}
  \Vc{c}_k =  \frac{F - k}{F} \, \Vc{c}_0 + \frac{k}{F} \, \Vc{c}_F.
\end{equation}
This is illustrated by the example in Fig.~\ref{fg:BTCDiag}, where $Q=2$. 

% ·  ·  ·  ·  ·  ·  ·  ·  ·  ·  ·  ·  ·  ·  ·  ·  ·  ·  ·  ·  ·  ·  ·  ·  ·  ·  ·  ·  ·  ·  ·  ·  ·  ·  ·  ·  ·  ·  ·  ·
% Block Truncation Coding diagrams 
%\begin{figure}[htbp]
% ·  ·  ·  ·  ·  ·  ·  ·  ·  ·  ·  ·  ·  ·  ·  ·  ·  ·  ·  ·  ·  ·  ·  ·  ·  ·  ·  ·  ·  ·  ·  ·  ·  ·  ·  ·  ·  ·  ·  ·

The pixels in a block are compressed using $M \times N \times Q$~bits that indicate the index of the best approximation within the reference line, i.e., for each pixel with RGB vector $\Vc{v}_{i,j}$ the reproduction vector index $k(i,j)$ corresponds to
\begin{equation}
  k(i,j) = \mathop{\operatorname{argmin}}_{k=0,1,\ldots,F} ||\Vc{v}_{i,j} - \Vc{c}_k ||,
\end{equation}
for $i = 0, 1, \ldots, M - 1,$ and $j = 0, 1, \ldots, N - 1$.

\begin{figure}
\centerline{\includegraphics[scale=0.7]{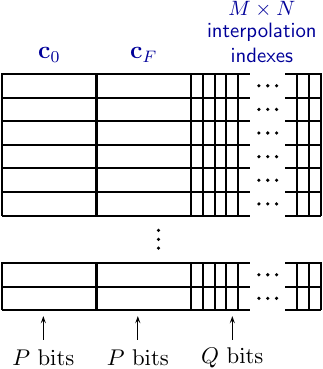}}
\caption{Data organization of BTC compressed blocks: $2 \times P$ bits are used for reference color vectors $c_0$ and $c_F$, and $M \times N \times Q$ bits for interpolation indexes\label{fg:BTCBits}.}
\end{figure}

Fig.~\ref{fg:BTCBits} shows how the BTC compressed data is organized in blocks with the same number of bits. The values of $P$ and $Q$ are commonly chosen so that the total number of bits used to represent a pixel block, $2 \times P + M \times N \times Q$, is a multiple of 8.

% ·  ·  ·  ·  ·  ·  ·  ·  ·  ·  ·  ·  ·  ·  ·  ·  ·  ·  ·  ·  ·  ·  ·  ·  ·  ·  ·  ·  ·  ·  ·  ·  ·  ·  ·  ·  ·  ·  ·  ·
% Block Truncation Coding bit arrangement 
%\begin{figure}[htbp]
% ·  ·  ·  ·  ·  ·  ·  ·  ·  ·  ·  ·  ·  ·  ·  ·  ·  ·  ·  ·  ·  ·  ·  ·  ·  ·  ·  ·  ·  ·  ·  ·  ·  ·  ·  ·  ·  ·  ·  ·

One important feature of this type of vector quantization is that all the information needed to decode a block is obtained from the compressed data, and there is no form of shared codebook. In fact, the interpolated colors are equivalent to VQ  codebooks, but are computed for each set of $M \times N$ pixels. This greatly improves performance  during parallel decoding, since it eliminates the need to copy data from large codebook memory.

%=======================================================================================================================
\subsection{Standard texture compression}
%=======================================================================================================================

Among the most used standards we have the Block Compression (BC) formats, numbered BC1, BC2, \ldots, BC7, that are supported by Microsoft’s Direct3D~11 specification~\cite{MSoft:25:tbc}, and Adaptive Scalable Texture Compression (ASTC)~\cite{Nystad:12:ast}, proposed by ARM and AMD, was standardized by the Khronos Group in 2013. Since those are used in many video game implementations, modern GPUs integrate specialized hardware to decompress data from those compression formats.

Those standards for graphics texture compression employ and extend BTC as described in the previous section. For instance, adding an alpha channel (transparency) for RGBA data, or the option to have two or three pairs of reference RGB vectors per block. While we can have blocks with different sizes, the most commonly used is $M=N=4$, i.e., they are applied to blocks of $4 \times 4$ RGB or RGBA vectors.

%=======================================================================================================================
\subsection{Clustering via color sorting}\label{sc:Cluster}
%=======================================================================================================================

% ·  ·  ·  ·  ·  ·  ·  ·  ·  ·  ·  ·  ·  ·  ·  ·  ·  ·  ·  ·  ·  ·  ·  ·  ·  ·  ·  ·  ·  ·  ·  ·  ·  ·  ·  ·  ·  ·  ·  ·
% Bit interleaving for generating 2D zigzag scan
%\begin{figure}[htbp]
\begin{figure}
\centerline{\includegraphics[scale=0.6]{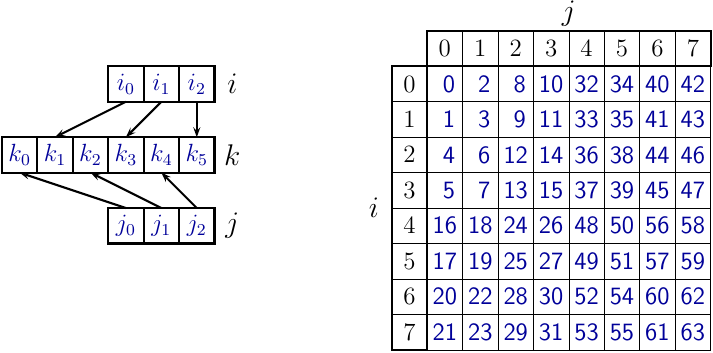}}
\caption{Example of two-dimensional recursive zigzag scan of an $8 \times 8$ block, generated by interleaving index bits.\label{fg:ZigZagScan}}
\end{figure}
% ·  ·  ·  ·  ·  ·  ·  ·  ·  ·  ·  ·  ·  ·  ·  ·  ·  ·  ·  ·  ·  ·  ·  ·  ·  ·  ·  ·  ·  ·  ·  ·  ·  ·  ·  ·  ·  ·  ·  ·

% ·  ·  ·  ·  ·  ·  ·  ·  ·  ·  ·  ·  ·  ·  ·  ·  ·  ·  ·  ·  ·  ·  ·  ·  ·  ·  ·  ·  ·  ·  ·  ·  ·  ·  ·  ·  ·  ·  ·  ·
% Bit interleaving for generating 3D color-ordered scans
%\begin{figure}[htbp]
\begin{figure}
\centerline{\includegraphics[scale=0.6]{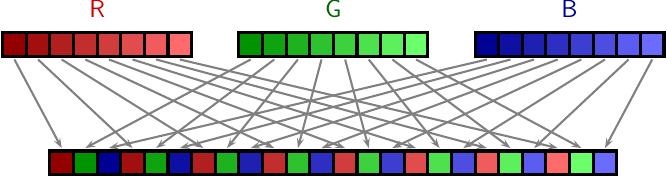}}
\caption{Generalization of the scheme in Fig.~\ref{fg:ZigZagScan}: interleaved bits in the bytes representing red, green, and blue generate 24-bit values that can be sorted, and the sorting order used for clustering color vectors.\label{fg:Bit4Scan}}
\end{figure}
% ·  ·  ·  ·  ·  ·  ·  ·  ·  ·  ·  ·  ·  ·  ·  ·  ·  ·  ·  ·  ·  ·  ·  ·  ·  ·  ·  ·  ·  ·  ·  ·  ·  ·  ·  ·  ·  ·  ·  ·

% ·  ·  ·  ·  ·  ·  ·  ·  ·  ·  ·  ·  ·  ·  ·  ·  ·  ·  ·  ·  ·  ·  ·  ·  ·  ·  ·  ·  ·  ·  ·  ·  ·  ·  ·  ·  ·  ·  ·  ·
% Image sorting examples
%
% Image 1: https://www.pexels.com/photo/fruit-on-towels-on-the-beach-19414557/
% Image 2: https://www.pexels.com/photo/exotic-fruits-in-market-20993671/
%
%\begin{figure}[htbp]
\begin{figure}
\centerline{\includegraphics[scale=0.6]{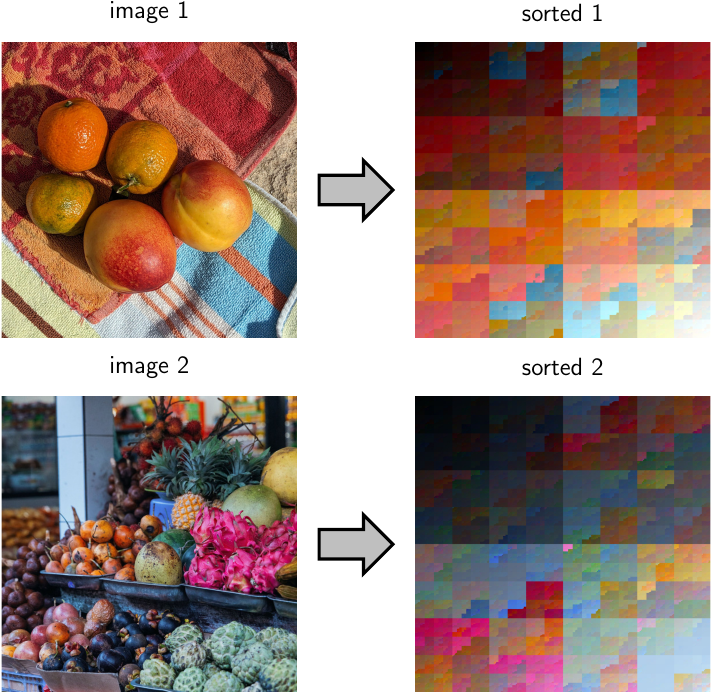}}
\caption{Examples of images that have pixels sorted according to values from RGB vectors, clustering similar color.\label{fg:ImgSort}}
\end{figure}
% ·  ·  ·  ·  ·  ·  ·  ·  ·  ·  ·  ·  ·  ·  ·  ·  ·  ·  ·  ·  ·  ·  ·  ·  ·  ·  ·  ·  ·  ·  ·  ·  ·  ·  ·  ·  ·  ·  ·  ·

As mentioned in the introduction, compression methods using BTC have the advantage of enabling random access and parallel decoding and rendering, but at the expense of not having the freedom to change the number of bits per block according to the block's properties.

While the number of bits used per BTC block cannot be changed, what varies per block is the reproduction error, and as we can observe from the example in Fig.~\ref{fg:BTCDiag} that this error depends on how the block's RGB vectors can be represented in a line segment. Blocks with large RGB vector variance cannot be well represented, while blocks that have colors highly clustered are much better reproduced. In fact, the reproduction error can go to zero if all RGB vectors are the same.

Next we show that we exploit this property to greatly improve the compression efficiency of BTC, if the grouping of RGB vectors for BTC coding can be modified so that those vectors are more similar.

There are many methods that can be used for reorganizing and clustering similar color vectors. To demonstrate that it possible to obtain good results without needing complex clustering, we employ a simple technique based on the well-known 2-dimensional recursive zigzag scans, which can be easily generated by interleaving bits from the 2-dimensional position indexes, as shown in Fig.~\ref{fg:ZigZagScan}.

The process shown in Fig.~\ref{fg:ZigZagScan} can be extended to 3-dimensional RGB vectors by using the bit-interleaving scheme illustrated in Fig.~\ref{fg:Bit4Scan}, and the following sequence is used to group vectors for BTC coding.
\begin{enumerate}
 \item For each RGB vector, interleave bits of the three bytes representing red, green and blue, using the scheme shown in Fig.~\ref{fg:Bit4Scan}, to create 24-bit integer values.
 \item Sort those values in increasing or decreasing order.
 \item Group RGB vectors for BTC encoding using the order obtained for values from interleaved bits.
\end{enumerate}

 %this can be exploited to BTC , while this property cannot be effective, in terms of total number of compressed data bits, for 2D graphics textures, it can greatly improve the compression efficiency 

Fig.~\ref{fg:ImgSort} shows examples of the results of this RGB sorting process, where for easier 2D visualization, the final order is shown using a recursive zigzag scan of the type shown in Fig.~\ref{fg:ZigZagScan}. We can observe that the images with sorted RGB values contain the same information, but they are organized by the sorting process to have much higher local similarity between colors.

% ·  ·  ·  ·  ·  ·  ·  ·  ·  ·  ·  ·  ·  ·  ·  ·  ·  ·  ·  ·  ·  ·  ·  ·  ·  ·  ·  ·  ·  ·  ·  ·  ·  ·  ·  ·  ·  ·  ·  ·
% PSNR of sorted image
%\begin{figure}[htbp]
\begin{figure}
\centerline{\includegraphics[scale=0.6]{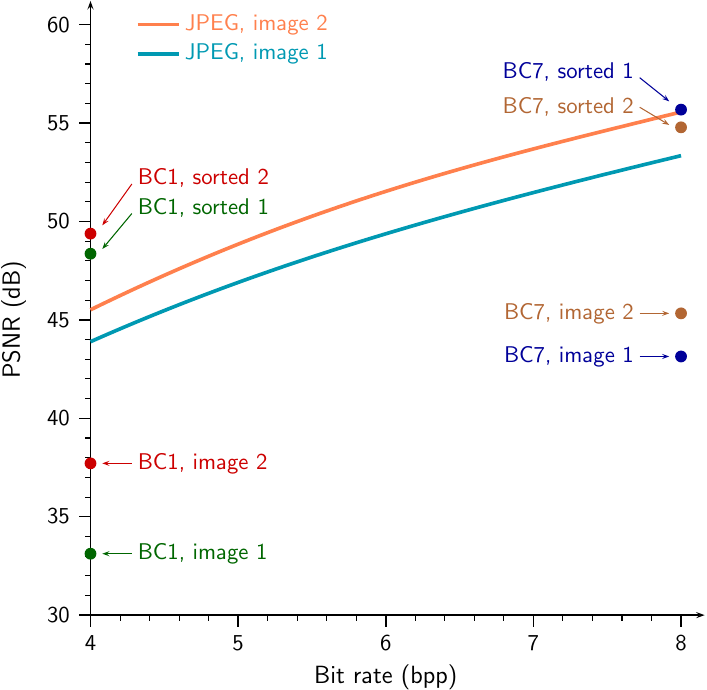}}
\caption{Comparison of JPEG, BC1, and BC7 compression on the images and their sorted versions shown in Fig.~\ref{fg:ImgSort}.\label{fg:SortedPSNR}}
\end{figure}
% ·  ·  ·  ·  ·  ·  ·  ·  ·  ·  ·  ·  ·  ·  ·  ·  ·  ·  ·  ·  ·  ·  ·  ·  ·  ·  ·  ·  ·  ·  ·  ·  ·  ·  ·  ·  ·  ·  ·  ·

Fig.~\ref{fg:SortedPSNR} shows how this greatly changes the effectiveness of BTC compression. We can observe that when the standard texture compression formats BC1 and BC7 are applied to the original images the results are much worse then those produced by JPEG, as measured by PSNR distortion. However, we can also observe that when those compression methods are applied to the images with sorted RGB vectors, the resulting PSNR jumps up by more than 10~dB, representing a very dramatic improvement in compression efficiency.

As explained in the introduction, this approach cannot be used for compressing 2-dimensional graphics textures because it would require extra bits for coding the indexes of each texels. 3DGS primitives, on the other hand, already have spatial positions as primitive data, so no extra data is needed after reordering. While it is better to also consider spatial positions during reordering, to keep primitives that are rendered together in nearby memory positions, BTC is applied to small blocks, so better compression can be also be achieved is color sorting is done independently in larger groups of primitives, so that spatial and color clustering can effectively work together. 

%<<<<<<<<<<<<<<<<<<<<<<<<<<<<<<<<<<<<<<<<<<<<<<<<<<<<<<<<<<<<<<<<<<<<<<<<<<<<<<<<<<<<<<<<<<<<<<<<<<<<<<<<<<<<<<<<<<<<<<<

%>>>>>>>>>>>>>>>>>>>>>>>>>>>>>>>>>>>>>>>>>>>>>>>>>>>>>>>>>>>>>>>>>>>>>>>>>>>>>>>>>>>>>>>>>>>>>>>>>>>>>>>>>>>>>>>>>>>>>>>
\section{Organizing 3DGS parameters for compression}\label{sc:Adapt}
%>>>>>>>>>>>>>>>>>>>>>>>>>>>>>>>>>>>>>>>>>>>>>>>>>>>>>>>>>>>>>>>>>>>>>>>>>>>>>>>>>>>>>>>>>>>>>>>>>>>>>>>>>>>>>>>>>>>>>>>

%=======================================================================================================================
\subsection{Hierarchical compressed data organization}
%=======================================================================================================================
	
The first aspect to be considered for developing practical compression for 3DGS is the necessity of having a compressed data format able to efficiently support large scale novel view synthesis~\cite{Meuleman:25:otf}, where scenes comprise a large area (e.g., a whole city block), and each part can be represented with millions of primitives.

In those large scenes the rendering process must efficiently cover the following range of viewing conditions.
\begin{enumerate}
	\item Views covering a small part of the scene must use a only the small fraction of primitives that define that part, ignoring the rest.
	\item Views covering large scene areas cannot render all the primitives contained in the view frustrum because the number is too high, but also because adding unfiltered details can cause very objectionable aliasing artifacts.
\end{enumerate}

Those are fundamental problems in computer graphics, and the solutions developed for this problem use hierarchical representations of the scene, which are used to change the rendering process to support a variable level of detail (LOD) according to the view distances.

\begin{figure}
\centerline{\includegraphics[scale=0.6]{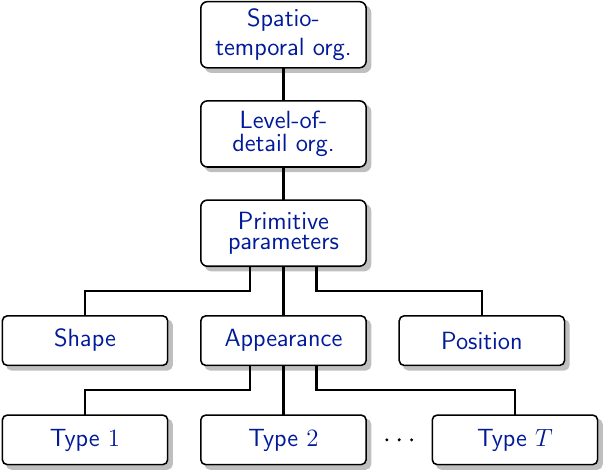}}
\caption{Hierarchical organization of compressed primitive parameters to enable efficient rendering of large and highly detailed scenes without aliasing artifacts and extended to support variable bitrate and random access.\label{fg:FileHier}}
\end{figure}

A format for compressed 3DGS scenes must also support that approach, using a hierarchical representation as shown in Fig.~\ref{fg:FileHier}. This is similar to other graphics representations, extended to support two additional requirements:
\begin{enumerate}
	\item The position and shape parameters are coded separately from appearance, since they are needed to determine if the primitive can be visible in current rendered frame (i.e., belongs to the view frustum).
	\item Appearance parameters can use different number of bits, and to enable random access, those with same number of bits are put together in different bitstreams.
\end{enumerate}
 
% ·  ·  ·  ·  ·  ·  ·  ·  ·  ·  ·  ·  ·  ·  ·  ·  ·  ·  ·  ·  ·  ·  ·  ·  ·  ·  ·  ·  ·  ·  ·  ·  ·  ·  ·  ·  ·  ·  ·  ·
% Compressed data file hierarchy
%\begin{figure}[htbp]
% ·  ·  ·  ·  ·  ·  ·  ·  ·  ·  ·  ·  ·  ·  ·  ·  ·  ·  ·  ·  ·  ·  ·  ·  ·  ·  ·  ·  ·  ·  ·  ·  ·  ·  ·  ·  ·  ·  ·  ·

\begin{figure}
\centerline{\includegraphics[scale=0.6]{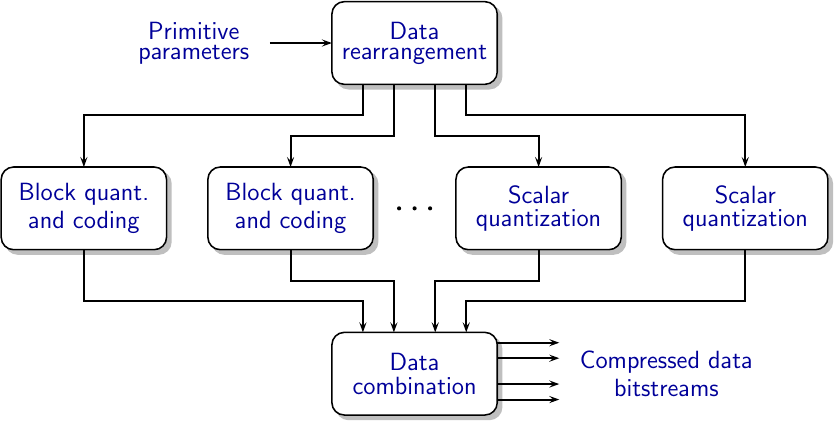}}
\caption{3DGS parameter separation for generating multiple compressed data bitstreams.\label{fg:DatSep}}
\end{figure}

Examples of coding methods generating different number of bits are described in the next section. Here the important observation is that primitives coded with different methods are separated as different types, as shown in Fig.~\ref{fg:FileHier}, which need to use different decoders before rendering. In general, the compressed data format uses the scheme shown in Fig.~\ref{fg:DatSep}, where primitives are first rearranged according to some parameter properties. For example, primitives with similar colors are put together to improve compression.

% ·  ·  ·  ·  ·  ·  ·  ·  ·  ·  ·  ·  ·  ·  ·  ·  ·  ·  ·  ·  ·  ·  ·  ·  ·  ·  ·  ·  ·  ·  ·  ·  ·  ·  ·  ·  ·  ·  ·  ·
% Primitive data seperation
%\begin{figure}[htbp]
% ·  ·  ·  ·  ·  ·  ·  ·  ·  ·  ·  ·  ·  ·  ·  ·  ·  ·  ·  ·  ·  ·  ·  ·  ·  ·  ·  ·  ·  ·  ·  ·  ·  ·  ·  ·  ·  ·  ·  ·

Next, different types of parameters are separated according to how they can be best quantized and encoded. For example, vector quantization may not be efficient for shape parameters, and instead they can be simply quantized to a fixed number of bits per parameter. The bits obtained from quantization are put together into different bitstreams so that the decoding method is defined by the index uniquely identifying a bitstream.

% ·  ·  ·  ·  ·  ·  ·  ·  ·  ·  ·  ·  ·  ·  ·  ·  ·  ·  ·  ·  ·  ·  ·  ·  ·  ·  ·  ·  ·  ·  ·  ·  ·  ·  ·  ·  ·  ·  ·  ·
% Parallel 3DGS decoding and rendering
%\begin{figure}[htbp]
\begin{figure}
\centerline{\includegraphics[scale=0.7]{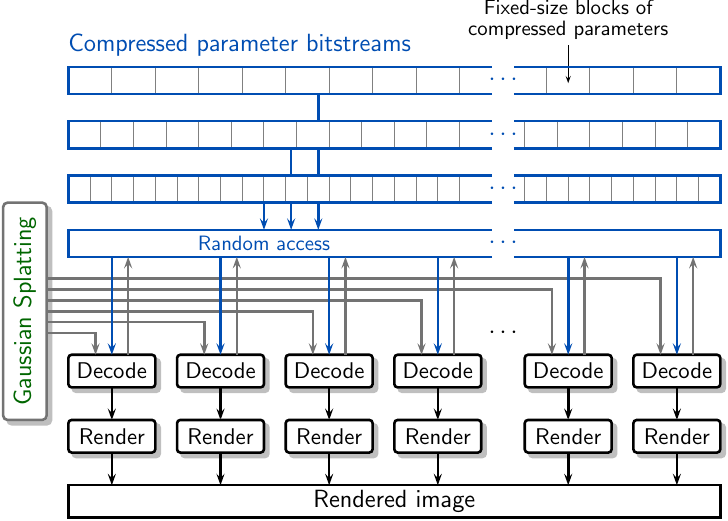}}
\caption{Parallel decoding of 3DGS parameters and rendering, enabled by random access to compressed data organized in different bitstreams, according to block organization and size.\label{fg:GSRend}}
\end{figure}
% ·  ·  ·  ·  ·  ·  ·  ·  ·  ·  ·  ·  ·  ·  ·  ·  ·  ·  ·  ·  ·  ·  ·  ·  ·  ·  ·  ·  ·  ·  ·  ·  ·  ·  ·  ·  ·  ·  ·  ·

%=======================================================================================================================
\subsection{Multiple bitstreams for variable bitrates and random access}
%=======================================================================================================================

As explained above, using the same number of bits to code a certain type of information is essential to enable random access for parallel decoding and rendering. This leads to suboptimal compression, since it is better to assign different number of bits depending on the entropy of the data. Graphics texture data is rigidly organized into two-dimensional arrays, so there are no practical ways to change the number of bits, except between different textures.

On the other hand, there is flexibility in rearranging 3DGS primitives, because their visual appearance is solely defined by their parameters. In fact, in the original rendering method, primitives must be sorted according to distance to camera in order to use the alpha blending of eq.~(\ref{eq:preblend}).

We exploit this property to improve compression, reorganizing parameters from a variable number of 3DGS primitives to data blocks that are compressed to a fixed number of bits. This way, we can vary the number of bits per primitive without changing the number of bits per block. To enable efficient random access, data from different types of arrangements are put together into different bitstream, as shown in Figure 6.

There is a great deal of flexibility in implementing this technique, since there is a very large number of ways to vary the number of primitives and the way their parameters are put together into blocks that are coded with a fixed number of bits.

For example, a straightforward way to compress the 16~RBG vectors with 3DGS spherical harmonic parameters is to reorganize them in $4 \times 4$ RGB blocks, as shown in Figure 8, and use a vector quantization method like Block Truncation Coding for compression.

\begin{figure*}
\centerline{\includegraphics[scale=0.7]{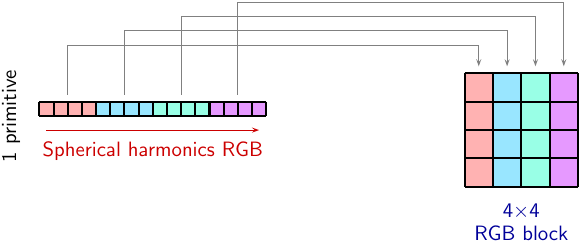}}
\caption{Rearrangement of 16~RGB vectors from one 3DGS primitive to one $4 \times 4$ RGB block.\label{fg:SetupA}}
\end{figure*}

Alternatively, for better compression we can exploit the similarity between the same type of parameters from different primitives by putting them together into blocks. For instance, it is possible to rearrange spherical harmonic RGB data from four primitives into four $4 \times 4$ RGB blocks, as shown in Fig.~\ref{fg:SetupA}.

% ·  ·  ·  ·  ·  ·  ·  ·  ·  ·  ·  ·  ·  ·  ·  ·  ·  ·  ·  ·  ·  ·  ·  ·  ·  ·  ·  ·  ·  ·  ·  ·  ·  ·  ·  ·  ·  ·  ·  ·
% Rearrangement example: 1 primitive -> 1 4x4 block
%\begin{figure}[htbp]
% ·  ·  ·  ·  ·  ·  ·  ·  ·  ·  ·  ·  ·  ·  ·  ·  ·  ·  ·  ·  ·  ·  ·  ·  ·  ·  ·  ·  ·  ·  ·  ·  ·  ·  ·  ·  ·  ·  ·  ·

\begin{figure*}
\centerline{\includegraphics[scale=0.7]{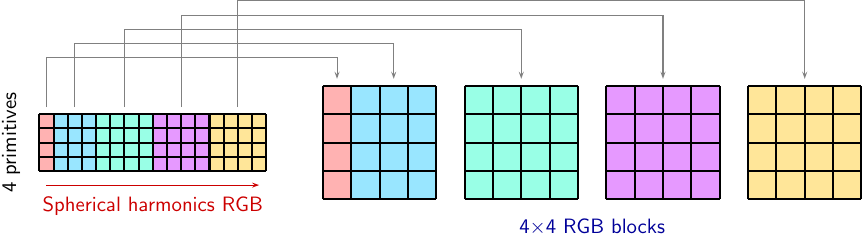}}
\caption{Rearrangement of 64~RGB vectors from four 3DGS primitives to four $4 \times 4$ RGB blocks.\label{fg:SetupB}}
\end{figure*}

While the number of bits per primitive used in the schemes of Fig.~\ref{fg:SetupA} and in Fig.~\ref{fg:SetupB} is the same, it is important to consider that the encoder can choose primitives with similar RGB vectors to share blocks such that, when using a vector quantization methods like BTC, reduces the reproduction error (compression distortion).

% ·  ·  ·  ·  ·  ·  ·  ·  ·  ·  ·  ·  ·  ·  ·  ·  ·  ·  ·  ·  ·  ·  ·  ·  ·  ·  ·  ·  ·  ·  ·  ·  ·  ·  ·  ·  ·  ·  ·  ·
% Rearrangement example: 4 primitives -> 4 4x4 blocks
%\begin{figure}[htbp]
% ·  ·  ·  ·  ·  ·  ·  ·  ·  ·  ·  ·  ·  ·  ·  ·  ·  ·  ·  ·  ·  ·  ·  ·  ·  ·  ·  ·  ·  ·  ·  ·  ·  ·  ·  ·  ·  ·  ·  ·

Another property that can be exploited is the fact that spherical harmonic coefficients are used to model view-dependent color, which may not be important in a large fraction of primitives. Commonly those parameters are represented with very small values, which can be converted to zero with little view quality degradation.

\begin{figure*}
\centerline{\includegraphics[scale=0.7]{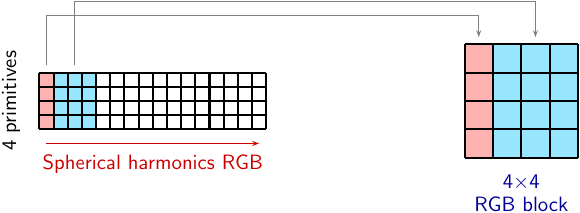}}
\caption{Rearrangement of 16~RGB vectors from four 3DGS primitives to one $4 \times 4$ RGB block, when higher order SH coefficients are all zeros.\label{fg:SetupC}}
\end{figure*}

\begin{figure*}
\centerline{\includegraphics[scale=0.7]{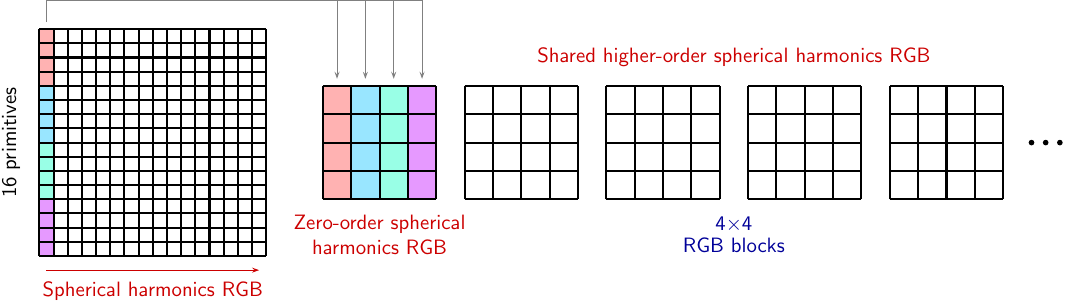}}
\caption{Rearrangement of 16~RGB vectors from 16 3DGS primitives to one or more $4 \times 4$ RGB blocks, depending on the number of nonzero blocks.\label{fg:SetupD}}
\end{figure*}

In those cases, we can define different bitstreams to primitives that use a reduced number of nonzero spherical harmonic RGB vectors. Fig.~\ref{fg:SetupC} shows an example of how a set of 16 RGB vectors from four primitives can be encoded using a single $4 \times 4$ RGB block, and Fig.~\ref{fg:SetupD} shows an example where single RGB vectors from 16 primitives are encoded using a $4 \times 4$ RGB block

% ·  ·  ·  ·  ·  ·  ·  ·  ·  ·  ·  ·  ·  ·  ·  ·  ·  ·  ·  ·  ·  ·  ·  ·  ·  ·  ·  ·  ·  ·  ·  ·  ·  ·  ·  ·  ·  ·  ·  ·
% Rearrangement example: 4 primitives -> 1 4x4 block
%\begin{figure}[htbp]
% ·  ·  ·  ·  ·  ·  ·  ·  ·  ·  ·  ·  ·  ·  ·  ·  ·  ·  ·  ·  ·  ·  ·  ·  ·  ·  ·  ·  ·  ·  ·  ·  ·  ·  ·  ·  ·  ·  ·  ·

% ·  ·  ·  ·  ·  ·  ·  ·  ·  ·  ·  ·  ·  ·  ·  ·  ·  ·  ·  ·  ·  ·  ·  ·  ·  ·  ·  ·  ·  ·  ·  ·  ·  ·  ·  ·  ·  ·  ·  ·
% Rearrangement example: 4 primitives -> 1 to 16 4x4 blocks
%\begin{figure}[htbp]
% ·  ·  ·  ·  ·  ·  ·  ·  ·  ·  ·  ·  ·  ·  ·  ·  ·  ·  ·  ·  ·  ·  ·  ·  ·  ·  ·  ·  ·  ·  ·  ·  ·  ·  ·  ·  ·  ·  ·  ·

The examples described above demonstrate that, by organizing data from multiple parameters, it is possible to improve compression by varying the number of bits per primitive, or better exploiting similarities between primitive parameters, even when using the same number of bits.

The $4 \times 4$ RGB blocks are commonly used in the BCx texture compression formats~\cite{MSoft:25:tbc}, but this type of SH coefficient rearrangement scheme can be extended to any form of compression based on Block Truncation Coding, or vector quantization.

%<<<<<<<<<<<<<<<<<<<<<<<<<<<<<<<<<<<<<<<<<<<<<<<<<<<<<<<<<<<<<<<<<<<<<<<<<<<<<<<<<<<<<<<<<<<<<<<<<<<<<<<<<<<<<<<<<<<<<<<

%>>>>>>>>>>>>>>>>>>>>>>>>>>>>>>>>>>>>>>>>>>>>>>>>>>>>>>>>>>>>>>>>>>>>>>>>>>>>>>>>>>>>>>>>>>>>>>>>>>>>>>>>>>>>>>>>>>>>>>>
\section{Experimental results}\label{sc:Result}
%>>>>>>>>>>>>>>>>>>>>>>>>>>>>>>>>>>>>>>>>>>>>>>>>>>>>>>>>>>>>>>>>>>>>>>>>>>>>>>>>>>>>>>>>>>>>>>>>>>>>>>>>>>>>>>>>>>>>>>>

%=======================================================================================================================
\subsection{Experimental setup}
%=======================================================================================================================

As explained in Section~\ref{sc:Param}, each 3GS primitive has 10~parameters representing opacity, position, and shape. In the experiments those were left uncompressed, and stored in 32-bit floating-point numbers. The texture compression was applied only to the spherical harmonic coefficients. This simplifies the analysis of results, to evaluate the effects of the RGB spherical harmonic coefficients compression with graphics texture coding methods.

Thus, data to be compressed corresponds to spherical harmonic coefficients $\tilde{\Vc{H}}$, that are pre-multiplied by primitive opacity as defined by eq.~(\ref{eq:premult}). To match the input image format of most texture compression methods, the first step before compression is to quantize floating-point values to 8-bit integer RGB components.

While the 3DGS learning process can be easily modified to produce values that regularly scaled for quantization and can be easily compressed, with minimal loss in reproduction quality, this is not used in the official 3DGS implementation~\cite{Kerbl:23:ofi}. Thus, some adjustments were needed to employ the well-known official 3DGS scene representations as reference.

Our solution was to first scan, for each scene, all the SH coefficients, and find the 16 scaling factors such that, after normalization, 99.5\% of coefficients are in the interval [-0.5,0.5]. The those normalized SH coefficients are scaled and quantized to integers between 0 and 255 for compression.

After quantization the SH coefficients were organized into $512 \times 512$ RGB images and compressed using the well-known {\tt compressonator} texture compression software~\cite{AMD:24:cmp} to generate files in dds format. Then images were decompressed with the same software, and their pixels were converted back to corresponding spherical harmonic coefficients. The BC1 and BC7 texture compression methods were used, with BC1 option {\tt -RefineSteps 2}, and BC7 option {\tt -Quality 0.2}. 

Two versions of each scene were created before quantization and compression. The first maintains the original sequence of primitives when generating images for texture compression. The second version has the primitives sorted according to the color of the first (diffuse) SH coefficient, using the method described in Section~\ref{sc:Cluster} and illustrated in Fig.~\ref{fg:ImgSort}.

Next, 3DGS scene views were rendered with the model using the decompressed data, and the resulting images were compared to the corresponding images rendered using the original 32-bit floating-point parameters, to obtain peak-signal-to-noise-ratios (PSNR). The final PSNR is the average over 25 views used for testing in~\cite{Kerbl:23:3dg}, which correspond to camera positions from the learning set.

%=======================================================================================================================
\subsection{Results and conclusions}
%=======================================================================================================================

Data from 3DGS scenes \textit{bicycle}, \textit{bonsai} and \textit{garden} was downloaded from the official 3DGS implementation site~\cite{Kerbl:23:ofi}, and used for all experiments. The PSNR results are shown in Table~\ref{tb:PSNR},
where the column \textit{SH bytes} represents the number of bytes used to encode the 48~SH coefficients per primitive. Note that BC1 compresses $4 \times 4$ RGB blocks to 8~bytes, and BC7 to 16~bytes. Different combinations using more than one primitive results in different averages per primitive.

% +  +  +  +  +  +  +  +  +  +  +  +  +  +  +  +  +  +  +  +  +  +  +  +  +  +  +  +  +  +  +  +  +  +  +  +  +  +  +  +
% Table: PSNR measurements
\begin{table*}
\centering
\caption{\label{tb:PSNR}Compression evaluation on selected 3DGS scenes.}
\begingroup
\renewcommand{\arraystretch}{1.2}
\begin{tabular}{|c|c|c|c|l|} \hline
 \multicolumn{3}{|c|}{Average PSNR (dB)} & SH & \multicolumn{1}{c|}{Method} \\ \cline{1-3} 
 Bicycle & Bonsai & Garden & bytes & \\ \hline \hline
 $\infty$ & $\infty$ & $\infty$ & 196 & 32-bit floating point \\
 46.20 & 50.17 & 48.92 & 48  & SH coeff. quantized to 8 bits \\ \hline
% ------------------------------------------------------------------------------------------------
 23.37 & 24.81 & 20.11 &  8  & BC1 on 16 SH coeff. packed in $4 \times 4$ blocks \\
 30.22 & 34.50 & 29.09 & 16  & BC7 on 16 SH coeff. packed in $4 \times 4$ blocks \\ \hline
% ------------------------------------------------------------------------------------------------
 30.73 & 31.88 & 29.86 & 10  & BC7 on 4 first SH coeff., BC1 on remaining (original) \\
 35.03 & 38.39 & 35.36 & 10  & BC7 on 4 first SH coeff., BC1 on remaining (sorted) \\ \hline
% ------------------------------------------------------------------------------------------------
 24.44 & 22.70 & 22.86 & 8.5 & BC1 on all $4 \times 4$ SH coeff. blocks (original) \\
 36.08 & 38.77 & 36.66 & 8.5 & BC1 on all $4 \times 4$ SH coeff. blocks (sorted) \\ \hline
% ------------------------------------------------------------------------------------------------
 31.53 & 29.84 & 29.34 &  9  & BC7 on $4 \times 4$ first SH coeff., BC1 on remaining (original) \\
 38.13 & 40.20 & 38.09 &  9  & BC7 on $4 \times 4$ first SH coeff., BC1 on remaining (sorted) \\ \hline
% ------------------------------------------------------------------------------------------------
 32.36 & 30.07 & 29.86 & 17  & BC7 on all $4 \times 4$ SH coeff. blocks (original) \\
 43.17 & 45.55 & 43.87 & 17  & BC7 on all $4 \times 4$ SH coeff. blocks (sorted) \\ \hline
\end{tabular}
\endgroup
\end{table*}
% +  +  +  +  +  +  +  +  +  +  +  +  +  +  +  +  +  +  +  +  +  +  +  +  +  +  +  +  +  +  +  +  +  +  +  +  +  +  +  +

We can observe in the second row of PSNR results in Table~\ref{tb:PSNR} that, even without changing the learning loss function, with proper scaling there is practically no visual loss when SH coefficients are only quantized to 8~bits.

The next two rows shows the results of using the simplest rearrangement scheme, shown in Fig.~\ref{fg:SetupA}, where all 16~SH coefficients are packed into $4 \times 4$ blocks. BC1 application results in significant quality degradation, and while results from BC7 are significantly better, the reproduction quality is still low.

This can be explained by the fact that the first SH coefficient, representing diffuse appearance, can be quite different from the remaining coefficient representing the specular effects, and thus cannot be well-represented with the BTC method discussed in Section~\ref{sc:BTC}.

The following two rows of results in were obtained using the rearrangement shown in Fig.~\ref{fg:SetupB}. There is a significant improvement in quality because BC7 uses 16~bytes per block, and also supports more than one quantization segment (reference vector pairs) per block, potentially using separate vector quantization for the diffuse and specular coefficients. In this arrangement we start to observe the advantages of sorting by color for BTC compression, resulting in 4.3, 6.5, and 5.5~dB PSNR improvements in the three tested scenes.

The last six rows of PSNR results in Table~\ref{tb:PSNR} were obtained using the rearrangement of Fig.~\ref{fg:SetupD}, employing 16 $4 \times 4$ RGB blocks, with each block containing SH coefficients of the same order from 16~3DGS primitives.

The first two of those rows show that the first SH coefficients (diffuse components) are very important for rendering quality, so that when they are coded together after color sorting PSNR improvements increase to 11.6, 16.1 and 13.8~dB, resulting in much better quality from BC1 compression.

This is confirmed with the results in the following two rows, that show that visually lossless compression can be achieved with slight increase in bytes per primitive by using BC7 in the first block and BC1 in the remaining 15. Finally, the last two rows show that, while using BC7 in all blocks yields the highest quality, it requires nearly doubling the number of bytes.

\begin{figure*}
\centerline{\includegraphics[scale=1.5]{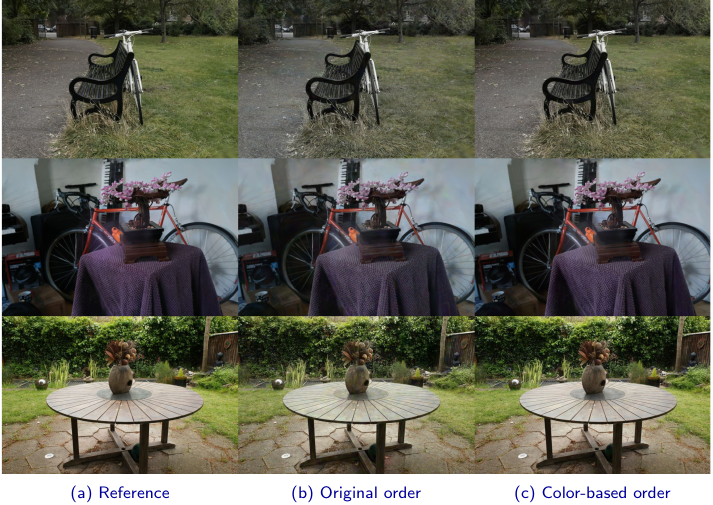}}
\caption{Comparisons of views generated from 3DGS scenes after compressing the 16 SH coefficients to 10~bytes per primitive. (a) Reference view rendered using the original 32-bit floating-point SH coefficients; (b) View rendered after decompressing the 16 SH coefficients to 10~bytes, without changing the order of primitives; (c) Same compression as (b), but rerdering SH coefficients according to RGB vectors.\label{fg:ResArray}}
\end{figure*}

Fig.~\ref{fg:ResArray} shows some examples of views rendered after using the proposed compression. The compression scheme used for those example is based on the rearrangement of Fig.~\ref{fg:SetupB}, with BC7 applied to the first block and BC1 to the remaining 3~blocks. It was chosen because its quality is sufficiently low to enable seeing some difference on side-by-side image comparisons. In other cases, it is much harder or impossible to visualize differences.

We can observe that the most visible compression-related distortion is in color appearance. However, since 3DGS primitives have very smooth boundaries, artifacts caused by color quantization do not create abrupt transitions, and are hard to identify. The same was observed in videos generated by moving camera position: all 3D features are preserved, and the most visible differences are in color.

% ·  ·  ·  ·  ·  ·  ·  ·  ·  ·  ·  ·  ·  ·  ·  ·  ·  ·  ·  ·  ·  ·  ·  ·  ·  ·  ·  ·  ·  ·  ·  ·  ·  ·  ·  ·  ·  ·  ·  ·
% 3x3 array of image views
%\begin{figure}[htbp]
% ·  ·  ·  ·  ·  ·  ·  ·  ·  ·  ·  ·  ·  ·  ·  ·  ·  ·  ·  ·  ·  ·  ·  ·  ·  ·  ·  ·  ·  ·  ·  ·  ·  ·  ·  ·  ·  ·  ·  ·

%<<<<<<<<<<<<<<<<<<<<<<<<<<<<<<<<<<<<<<<<<<<<<<<<<<<<<<<<<<<<<<<<<<<<<<<<<<<<<<<<<<<<<<<<<<<<<<<<<<<<<<<<<<<<<<<<<<<<<<<

%%%%%%%%%%%%%%%%%%%%%%%%%%%%%%%%%%%%%%%%%%%%%%%%%%%%%%%%%%%%%%%%%%%%%%%%%%%%%%%%%%%%%%%%%%%%%%%%%%%%%%%%%%%%%%%%%%%%%%%%
%\section{Section}\label{sc:}
%%%%%%%%%%%%%%%%%%%%%%%%%%%%%%%%%%%%%%%%%%%%%%%%%%%%%%%%%%%%%%%%%%%%%%%%%%%%%%%%%%%%%%%%%%%%%%%%%%%%%%%%%%%%%%%%%%%%%%%%

%=======================================================================================================================
%\subsection{Subsection}
%=======================================================================================================================

%-----------------------------------------------------------------------------------------------------------------------
%\subsubsection{Subsubsection}
%-----------------------------------------------------------------------------------------------------------------------

%>>>>>>>>>>>>>>>>>>>>>>>>>>>>>>>>>>>>>>>>>>>>>>>>>>>>>>>>>>>>>>>>>>>>>>>>>>>>>>>>>>>>>>>>>>>>>>>>>>>>>>>>>>>>>>>>>>>>>>>
\bibliographystyle{IEEEbib}
\bibliography{main}
%<<<<<<<<<<<<<<<<<<<<<<<<<<<<<<<<<<<<<<<<<<<<<<<<<<<<<<<<<<<<<<<<<<<<<<<<<<<<<<<<<<<<<<<<<<<<<<<<<<<<<<<<<<<<<<<<<<<<<<<

%<<<<<<<<<<<<<<<<<<<<<<<<<<<<<<<<<<<<<<<<<<<<<<<<<<<<<<<<<<<<<<<<<<<<<<<<<<<<<<<<<<<<<<<<<<<<<<<<<<<<<<<<<<<<<<<<<<<<<<<
\end{document}